# MLOps: A Primer for Policymakers on a New Frontier in Machine Learning

## An Explainer of Tools for Bias Mitigation in the MLOps Lifecycle

**Jazmia Henry**
**July 18, 2022**

**Summary**

Discussions about reducing the bias present in algorithms have been on the rise since the mid 2010s. AI ethicists, DEI practitioners, Sociologists, Data Scientists and Social Justice Advocates have decried the lack of understanding of the harms that algorithms pose to people who belong to historically marginalized groups. These cries have become increasingly accepted in industry since 2020, but little is understood of how algorithm and Machine Learning (ML) model builders should go about mitigating bias in models that are intended for deployment.

This chapter is written with the Data Scientist or MLOps professional in mind but can be used as a resource for policy makers, reformists, AI Ethicists, sociologists, and others interested in finding methods that help reduce bias in algorithms. I will take a deployment centered approach with the assumption that the professionals reading this work have already read the amazing work on the implications of algorithms on historically marginalized groups by Gebru, Buolamwini, Benjamin and Shane to name a few. If you have not read those works, I refer you to the "Important Reading for Ethical Model Building " list at the end of this paper as it will help give you a framework on how to think about Machine Learning models more holistically taking into account their effect on marginalized people. In the Introduction to this chapter, I root the significance of their work in real world examples of what happens when models are deployed without transparent data collected for the training process and are deployed without the practitioners paying special attention to what happens to models that adapt to exploit gaps between their training environment and the real world.  The rest of this chapter builds on the work of the aforementioned researchers and discusses the reality of models performing post production and details ways ML practitioners can identify bias using tools during the MLOps lifecycle to mitigate bias that may be introduced to models in the real world.

**Introduction**

"Whether AI will help us reach our aspirations or reinforce the unjust inequalities is ultimately up to us." - Joy Buolowini, 'Facing the Coded Gaze' AI: More than Human[1]

Whether you're driving your car using a GPS system, call on Alexa or Siri to turn on your favorite tune, go on social media to perform a well-earned scroll down memory lane, or go to Google search to find a gift to buy for a friend, you have encountered a Machine Learning model. These models collect data to make predictions that can make our lives easier, or harder, depending on the care taken during the Machine Learning life cycle process. The Machine Learning life cycle refers to the multistep process that begins with understanding project objectives and ends at model deployment and maintainability. Each step in the process is as follows: define project objectives, acquire and explore data, model data, interpret and communicate, and implement, document, deploy and maintain the model.[2] The reason it is referred to as a life cycle is simple: at each step of the process, the ML practitioner is likely to go back a step or two to validate results by adjusting original project objectives, retrieving more or different data, tuning the model, or documenting their process for model maintenance. In AI ethics, researchers focus heavily on the first three steps in the process to find ways to mitigate bias downstream. This work focuses on the last step of the life cycle: implementing, documenting, and maintaining the model. This step is also referred to as Machine Learning

---

[1] Barbican Centre. "Joy Buolamwini: examining racial and gender bias in facial analysis software".

[2] n.a., "Machine Learning Lifecycle."

Operations or MLOps as referred to in this paper. Before getting into my argument as to why a focus on MLOps is an important addition to the AI ethical lexicon, I will first discuss past work and work my way into the importance of bias mitigation in the MLOps step of the ML lifecycle.

Batya Friedman, Peter Kahn Jr. and Alan Borning take a particular interest in part 1 of the cycle, defining project objectives, in their article, "Value Sensitive Design: Theories and Methods".[3] Within the article, Friedman et al propose a tripartite method to designing a system keeping human values in mind. This system can be an application that makes decisions that affect the public and for the purposes of this paper, readers can think of Machine Learning as a type of system. The tripartite method considers the conceptual, empirical, and technical components of a system's design to formulate its main objectives and the way practitioners should just model success. Conceptual considerations involve the values that are being supported or perpetuated in the model's build. The empirical method considers the tradeoffs of what can often be competing values and seeks to understand the desired outcome of the model. Here, practitioners should consider the difference between intent and impact. Just because a model intends to do good does not mean it will and if it commits harm then the impact should outweigh the intent of the model builder and the decision makers of the model must go back to the drawing board for realignment. The last method, technological investigations, looks into how technologies have been used in the past and if these technologies "have hindered or supported human values".[4]

In both *Algorithms of Oppression* by Safiya Noble and *Weapons of Math Destruction* by Cathy O'Neil, they perform a technological investigation and examine how technologies have

---

[3] Friedman, Kahn Jr., and Borning, "Value Sensitive Design: Theory and Methods."

[4] Friedman, Kahn Jr., and Borning, "Value Sensitive Design: Theory and Methods," 3,

been used to perpetuate harms in the past. They investigate where efforts to automate processes have been placed and question the validity of model use leaning on the side of using more simple technical solutions over ones that learn, adapt, and exploit biases.[5] [6] Noble goes so far as to suggest that a model's values have been, at its core, biased and these biases are amplified in the overall technology by design.[7] (Noble)

Joy Buolawini, Timnit Gebru, and Ruha Benjamin have work that focuses on part 2 of the Machine Learning lifecycle- acquire and explore data. Opposed to focusing on what technologies have done in the past, these researchers take what Andrew Ng refers to as the "data-centric" approach to close the gap between model performance and model values.[8] Their work comes after an instance of bias showing up in a model built by Google to label people and animals.

On June 29, 2015, Twitter user, Jacky Alciné, tweeted "Google Photos, y'all [expletive] up. My friend's not a gorilla."[9] (@jackyalcine) The post went viral with people around the world criticizing Google's latest classification algorithm.[10] (Wired) Scholarship on Algorithmic fairness soon followed with AI ethicists and researchers from around the world highlighting bias perpetuated by Machine Learning models. Joy Boulawini and Timnit Gebru releasing the article "Gender Shades: Intersectional Accuracy Disparities in Commercial Gender Classification" exposing the misclassification of Black women in large classification models. They set out to supplement the lack of image data of Black women using their own facial analysis dataset and incorporate it into three commercial classification models, IBM, Microsoft, and Face++. What they found was an outsized error rate when classifying Black women compared to Lighter completed men (8% compared to 34.7%). While supplementation improved performance on Black women, the classifier was still a long way from being equitable. Boulawini and Gebru conclude their work by advocating for transparency and accountability through "inclusive benchmark datasets and subgroup accuracy reports".[11]

---

[5] Noble, *Algorithms of oppression: How search engines reinforce racism*.

[6] O'Neil, *Weapons of Math Destruction: How Big Data Increases Inequality and Threatens Democracy*.

[7] Noble, *Algorithms of oppression: How search engines reinforce racism*.

[8] Ng, "MLOps."

[9] Acline, Jacky [@jackyalcine], 29 June 2015, Tweet.

[10] Simonite, "When It Comes to Gorillas."

In Race After Technology, Ruha Benjamin suggests a similar approach as the one prescribed by Gebru and Buolawini, increasing transparency using, what she calls, a Dataset Nutrition Label. Much like a Nutrition Label on the back of packaged food, the Dataset Nutrition Label will describe the contents of each model and their performance for tasks across subgroups.[12] If Google would have had a Dataset Nutrition Label and been more transparent in its Data collection process, solutions to the Google misclassification problem may have been found more quickly. Unfortunately, instead of finding a solution to the problem, three years later, Google simply deleted "Gorilla" and "Chimpanzee" tags from the offending model completely.[13] (Wired) While Dataset transparency is an important step in mitigating bias in the ML lifecycle, it forgets a very important reality of deploying AI in the real world: AI, by its nature, is built to adapt to new situations and, many times, can cover its biased assumptions creating far reaching problems down the line. This takes us into a common tension in model building observed by ML practitioners in industry: the difference between the work done in part 3 of the life cycle, model building, and part 5 of the life cycle, deploying and maintaining the model.

---

[11] Buolamwini and Gebru, "Gender Shades: Intersectional Accuracy Disparities in Commercial Gender Classification," in *FAT*, 2018.

[12] Benjamin, *Race after Technology: Abolitionist Tools for the New Jim Code*

[13] Simonite, "When It Comes to Gorillas."

An example of this is discussed in *You Look Like A Thing And I Love You* by Janelle Shane. Within it, she gives the example of AI being trained in a "dream world" with the skills acquired there transferred to the "real world".[14] Unfortunately, the rules that bind us in reality are not present in the "dream world" and the AI was able to exploit the glitches it learned in the dream world and tried to apply this in the real world. This caused the model to subsequently fail.[15] If you change the term "dream world" and replace it with "training environment", you quickly see how models can perpetuate biased reasoning even with balanced training data. AI can adapt to cover its mistakes by learning to exploit gaps in the training set for higher reward in the real world.

Microsoft learned this first hand in 2016 when they unveiled "Tay" - a conversational AI bot deployed on Twitter. Made to mine public data and incorporate the data it gathers from conversations with people online to become more conversational, the chatbot quickly picked up racist and homophobic attitudes even going so far as to praise Hitler in one of its Tweets within 24 hours of being online.[16] Microsoft never trained its bot on such divisive language. The bot was trained in a dream world where the interactions it had and the data it collected was upbeat and positive. Once deployed, it was no longer within the confines of its dreamlike training environment and like a new bowler bowling without guardrails, the bot's language quickly went into the gutter. While the details of the bots training environment was not made public, the bot seemed to be rewarded through its learning of new ways of interacting with human respondents. This led to the bot exploiting a gap between its training and what it was observing in the real world to gain maximal reward by engaging in divisive behavior- behavior that, undoubtedly, had more reactions online than positive ones based on the popularity it gained after sending hateful Tweets.[17] Dataset Nutrition Labels and Data transparency may have helped in Google's case, but it would not have helped fix Tay. Tay would have needed tools put in place during the deployment step of ML lifecycle management to ensure that poor behavior is not exploited for the model to receive greater reward.

---

[14] Shane, *You Look Like a Thing*.

[15] Shane, *You Look Like a Thing*.

[16] Price, "Microsoft Is Deleting Its AI Chatbot's Incredibly Racist Tweets," Business Insider.

    MLOps is the final line of defense between the model and the user interacting with the model. MLOps covers the last step in the ML lifecycle: implementation, deployment, and model maintenance.[18] By creating a path to bias mitigation within the MLOps function of the ML lifecycle, we can better ensure that the work that has been done to mitigate bias in parts 1 - 3 of the lifecycle are fortified and continued in the real world through a robust framework created with the last step of the ML lifecycle prioritized.

**Models in the Wild, Wild West**

---

[17] James Vincent, "Twitter Taught Microsoft's Friendly AI Chatbot to Be a Racist Asshole in Less than a Day," The Verge.

[18] n, a. "Machine Learning Life Cycle."

Algorithms work very differently once they have been deployed than they do on a practitioner's notebook or local machine. Veterans refer to models that have been deployed into production as models "in the wild", because once a model has been deployed and users are able to interact with it, it behaves much differently than it did when going through training. As well, the practitioner who built the model may have underestimated a sample and has much less control over exogenous variables that may affect the model's performance. Because of this, bias mitigation techniques that ignore the deployment portion of the Machine Learning lifecycle are at risk of augmenting bias in the model once it has been put into production. There are many strategies that have been introduced to get models "in the wild" under control with a specific emphasis on practitioners monitoring deployed models for concept or data drift in traditional MLOps methods. This method can be extended as a bias mitigation technique in the model deployment.

### *Concept and Data Drift as Clues*

Concept and data drift can both cover biases showing up in a model that has been deployed and should be looked at not only from a perspective of model performance but also of bias mitigation. Data drift occurs when the distribution of some input within the model is different in the wild than during model training. Concept drift occurs when the relationship between the model's input and output are different in model training than when deployed. Both concept and data drift can occur when a model has biased results against a group. This can hold true even if a practitioner goes above and beyond to mitigate bias in model building. It is important to understand the fundamentals of how this might happen before discussing its implications and how it can be addressed.

A simple example develops the point. Imagine a model is trying to identify people with comorbidities that took the COVID vaccine last year, and tries to predict if a person belongs to one of two populations within a relatively wealthy coastal city in the US. In this case, each population is decided by the labels "does not have a comorbidity" or "has a comorbidity". Let's say population *a ( does not have comorbidity)* is 95% of a model's training dataset and population *b ( has comorbidity)* is 5% of the training dataset. A Data Scientist aware of the imbalance in the dataset decides to run SMOTE, a method to equalize extremely imbalanced data by synthetically overestimating the minority population to match that of the major population. Once that is done, the Data Scientist trains the newly balanced dataset with the knowledge that doing so may result in a reduction in model performance, but they want to prioritize equal observations over maximized performance. After model deployment, however, the ML practitioner finds that the population of people interacting the model have a different distribution. Population *a* accounts for 89% of interacting users and population *b* accounts for 11%.

To be clear, depending on the overall size of the population and the targeted confidence interval, this change in percentage of the population may not be noticeable enough for a practitioner to change it. Every modeler has some margin of error that they deem acceptable once a model is in production as long as the model does not go outside of those bounds. On its face, this may not appear to be something of concern, but for practitioners tasked with model deployment, taking a closer look can glean some important results.

While Data Drift has been identified in the aforementioned model by the change in model input, practitioners benefit from looking at the distribution within each label and the features shared by each label and judge the variance of each population within the sample to establish true model confidence. If training data population *a* did a good job at generalizing to predict "in the wild" population *a* but did a poor job of predicting population *b,* one should stratify the sample to understand if there is variance between groups. Here is where demographic data can be used to overcome bias by allowing a ML practitioner to break the data into groups and examining the variance in performance between them. I will discuss this technique within the Statistical Methods in Model Monitoring section.

      Now, let's take the same example above, but instead of the proportion of the population's distribution in deployment differing from the training dataset, let's assume that the reason for the outcome is misunderstood in the dataset so the model is unable to adapt. Say there is a campaign in the area to get more people with comorbidities to get vaccinated. This causes population *a* to increase in training data share and population *b* to decrease. Not only this, but say, due to a lack of resources, people from less affluent areas with comorbidities come to the more affluent area to get vaccinated causing population a to decrease and population b's share to increase. During the model interpretation process, practitioners unaware of the campaign believe that people in less affluent areas with comorbidities are less likely to get vaccinated than people with comorbidities in affluent areas. The model they created then uses the share of the population with comorbidities that are getting the vaccine to allocate more resources to the affluent area causing the cycle of people with comorbidities from less affluent areas to get the vaccine in the more wealthy town to continue. Assumptions being made about this group based on model prediction may lend itself to bias simply because of this exogenous effect. This would be an example of Concept Drift occurring due to an introduction of bias in a model.

***Identify the Source of the Problem***

When Concept or Data Drift occurs, it is important for practitioners to make adjustments to the model through the strategic collection of more data. There are many ways that practitioners can search for where to begin to find more data and what sources of data would be most reliable. There are two main avenues to consider when trying to identify how to get to the bottom of this problem: 1) Internal data leakage and 2) External variable capture.

Internal data leakage is when the data source, data pipeline, query, model, dashboard, or other data based process has broken down. There are many stages at which this can happen and show up in model output as concept drift so be sure to thoroughly examine each one. Unless you have reason to believe otherwise, looking for signs of internal data leakage is the best place to start as this is normally the most likely cause of changes in model performance and is a much easier thing to prove and rectify. The second avenue to consider is external variable capture. This was shown in the example above- when an exogenous variable affects model performance, it is up to the practitioner to find data that supports the premise of an exogenous effect.

**Statistical Methods in Model Monitoring**

There are more techniques the ML practitioner can take advantage of beyond monitoring for Concept and Data Drift that use Statistical Analysis.

*Stratified sampling*

In the setup of a Model monitoring dashboard, ML practitioners could add a plot monitoring the scores of each group within a sample through a process called Stratified Sampling. Stratified sampling is defined by taking an overall sample and dividing it into subgroups for further analysis. In the case of making predictions on any population of people, creating subgroups around demographic factors can, at best, help lead to overall parity between groups and, at least, can allow business leaders and practitioners to see the difference in model performance across groups for further analysis downstream.

Here's what it would look like in practice. Consider an example of a Data practitioner monitoring performance of a model that has been deployed on a simple prediction model that is attempting to predict who would be more or less likely to place a bet. The labels are divided into two populations: population *a* which does not place bets and population *b* which does place bets. Within each population is a distribution of subgroups according to demographic. Perhaps, during model training, the Data Scientist notices that gender is a strong predicting factor of who would be likely to place bets so the ML practitioner monitoring the model may stratify model output by gender to see if there is a divergence of scores by that feature. This can allow observers to gain extra confidence in the model as strong features may cause model performance to be favored in the direction of one subgroup over another causing overall scores to be an incomplete picture of true model performance.

*Imbalanced Classification/ Model Weighting*

The idea of using imbalanced classification techniques in order to improve classification models is not new. In fact, Data Scientists frequently use imbalanced classification techniques when these problems arrive. Popular techniques such as SMOTE and Near-Miss are frequently used to improve performance in cases of predictive algorithms. For the purposes of this paper I will focus on using imbalance classification techniques to assist with instances of bias in model training and model performance.

When working on a machine learning model that is meant to be used by the general population, it is important for the data scientist to identify ways in which some groups may have biased results over others. This can be done with the stratified sampling technique I discussed above. Once they have identified that there is an imbalance within the group's, however, they have the ability to use demographics as a label and then apply SMOTE or Near-Miss (or some other imbalanced classification method) to increase model performance for the underrepresented group.

*Experiments and Analytics*

In the case of experimentation, it is very important for the ML practitioner to make sure that they separate out the experimentation groups based on sub groups. After they have done so it is important to judge the efficiency of an experiment not simply based on the mean or median scoring, but over the overall performance of each subgroup.

This is where the data analyst and the ML practitioner can collaborate on the best way of gauging experimentation performance optimality. The data analyst can then analyze the results coming from the experiment to use as extra data that can be fed back into the model in future versioning and model tuning.

Experimentation is an important tool in the ML practitioner's toolbox. Experiments can be used to prove model efficiency in the wild, to compare performance of multiple models applied to the same problem, or even be used to prove if a model is a superior solution over a simple technical solution. Practitioners should take care to separate users into treatment and control groups, stratify each group into subgroups based on historically marginalized groups, and compare the performance of the ML model when serving predictions to people versus a simple

Tech solution that gives people offerings based on explicit user selections. The practitioners should then monitor observations and judge model efficiency on how well it performs on marginalized groups compared to the simple Tech solution.

*Model Parity Using Confusion Matrix*

Model parity is a technique commonly discussed during discussions about fairness in Machine Learning models amongst circles of ethicists and policy makers. There are different types of parity metrics that can be used to mitigate bias in a model and modeling dashboard for machine learning practitioners to monitor model performance.

The first type of parity metric we are going to discuss is one that is quite familiar to most: demographic parity. Demographic parity requires that the positive rates of the underrepresented subgroup be equal to the percentage of the positive rate of the over-represented class. The probability equation of the outcome is defined as :

$$P(\hat{Y} \mid A = 0) = P(\hat{Y} \mid A = 1)$$

Such that the Positive Rate of the over-represented subgroup a = 0 is the same as the Positive Rate of under-represented subgroup a = 1. A ML practitioner can simply use a Confusion Matrix to demonstrate if each subgroup has the same Positive Rate.

The second type of parity is called the Equal Opportunity parity metric. This parity metric takes things a step further and says that, assuming all things are equal, the under-represented subgroup should receive some prediction at the same rate as the over-represented group. In this case, the ML practitioner would use the True Positive Rate within the Confusion

Matrix for each subgroup to identify if the outcomes are equal. The formula of this probability is as followed:

$$P(\hat{Y} = 1 \mid A = 0, Y = 1) = P(\hat{Y} = 1 \mid A = 1, Y = 1)$$

Such that the True Positive Rate of the over-represented subgroup a = 0 is the same as the Positive Rate of under-represented subgroup a = 1.

The last type of parity I will discuss within this paper is the Equalized Odds parity metric. This metric is one that not only suggests that outcomes should be equal suggesting known factors remain constant but also that False Positives across subgroups must be equal. To get the Equalized Odds parity rate, the ML practitioner would need to divide on the True Positive Rate of each subgroup by the False Positive Rate for each subgroup. The goal is to make the percentage of the Equalized Odds parity metric as close to zero as possible. The probability formula is below:

$$P(\hat{Y} = 1 \mid A = 0, Y = y) = P(\hat{Y} = 1 \mid A = 1, Y = y), y \in \{0, 1\}$$

Such that the True Positive Rate of the over-represented group a = 0 is equal to the True Positive Rate of the under-represented subgroup a = 1 **and** the False Positive Rate of the over-represented group a=0 is equal to the False Positive Rate of the under-represented subgroup a = 1.

**Deployment Methods for Bias Mitigation**

Of all of the deployment methods in a Machine Learning Operative's arsenal, there are two that can be used specifically for the purpose of bias mitigation that I will talk about in this paper. The first one is Canary deployment and the second is Blue / Green deployment.

*Canary Deployment*

Canary deployment is so aptly named based on an old mining strategy. If you ever heard of the "Canary in the coal mine" saying then you are somewhat familiar with the heart of this process. When coal miners would approach a new mine, they would release a canary into the coal mine to test if harmful gasses were present. If the canary was able to survive entering and leaving the coal mine then the coal miners would feel assured that the coal mine was safe to be entered. If not, then the coal miners would go back to step one and apply methods to reduce the toxic gasses until the coal mine is safe to enter.

Bias in a model is a type of toxic gas. When a model is making predictions that unjustly discriminates against a historically marginalized group it is to the detriment of the performance of the model, and, in some cases, to society as a whole. When the ML practitioner uses Canary deployment they have the ability to test if model performance degradation has occurred as a result of bias that has been introduced to the model "in the wild".

This process can be used after stratified sampling techniques that separate a population set into subgroups based on demographics and/or analyzing the variance between said subgroups. The ML practitioner can schedule a staggered release of the model to an equitable sample of people in the entire population through the process of Canary deployment. In practice, this will look like taking a representative sample of some percentage $n$ of a population and unveiling the model to this group over some $T$ time and testing the interactions of the model across each group while analyzing model parity. If the model performs up to an acceptable standard, then the ML practitioner can safely extend the deployed model to a larger population *n + 1* until model stability has been achieved.

*Blue/Green Deployment*

The second deployment strategy is called Blue / Green deployment. This is a process where the ML practitioner can slowly move people from one production environment to another. Here, applications of experimentation can be employed for further model testing. ML practitioners can test the efficiency of model 1 versus model 2 and examine how much bias is present in each model as a KPI then move forward with the model that has the best parity performance. They can also compare the status quo to the introduction of a new machine learning production environment and compare parity of the model over time. Either way, this method allows for a ML practitioner to compare the efficiency of model performance through the lens of bias mitigation. They can then work with their team after some *T* time to examine model efficiency.

*Human in The Loop*

A wonderful way to reduce bias either picked up by a model in the wild or covered up in model training by exploitative decision making by the model is returning observations picked up by the model for retraining purposes to a database for human approval. This can be done in a couple ways:

**Flagging.** ML practitioners can initiate a flagging mechanism that causes the model to err on the side of human deferment. This can be done by tracking scoring per observation in a database that tracks predicted observations and marking predicted observations that do not score high enough. The definition of adequate scoring should be based on the originating error rate identified during the training process and tracked in the Dataset Nutrition Label. So, for example, if a classifier has a F1 score of 88% for Black Women within the training environment, based on overall variance between training and deployed performance, practitioners may decide to flag all observations with F1 scores outside of the range of acceptability and find a questionable score to be 84% or lower. These observations can be flagged for further review by practitioners and logged for model transparency.

**Pruning.** Pruning is when a practitioner removes an observation for model retraining. Practitioners can decide to do this as a way of erecting guardrails that mitigate bias being introduced by the model exploiting a gap between its training environment and the real world.

**Nudging.** Observations flagged in the retraining set can be returned to the human for the human to hardcode the appropriate response for the model to learn from.

**Conclusion**

What would have happened to Microsoft if, instead of simply releasing Tay without taking proper care to mitigate bias during the MLOps cycle, they had employed some of the techniques discussed in this chapter? Using Concept Drift, practitioners would begin to notice that the environment that Tay had been deployed into was drastically different from the environment they trained her in. They would begin to see a preponderance of data being ingested that was not present in their training environment using Data drift and could have adapted accordingly. Using a canary deployment approach, they could have tested Tay in a real world environment with fewer people and segmented the people using Tay into test and control groups using experimentation. They could have compared Tay's performance as a conversational AI bot to a bot with hardcoded responses and seen how well each bot performed in the real world.

Lastly, practitioners could have flagged new words introduced into Tay's training lexicon, removed profanity and divisive words using pruning, and nudged the dataset by replacing the inappropriate words with more appropriate ones.

From Defining Project Objectives to Model Deployment, every step of the ML lifecycle should include bias mitigation techniques. If we focus on steps 1-3 and do not take care to understand step 5 and how it contributes to bias as well, then we will not be able to have the equitable and value driven models that are good for society. While this paper serves as an overview of a few techniques that Machine Learning practitioners can use to mitigate bias in a model during the MLOps LifeCycle, there are more techniques that can and should be explored by people passionate about making Machine Learning models more equitable for those who have been historically marginalized.